\documentclass[11pt,letterpaper]{article}
\usepackage{emnlp2016}
\usepackage{times}
\usepackage{latexsym}
\usepackage{graphicx}
\usepackage{tabularx}
\usepackage{amsmath}
\usepackage{url}
\usepackage{multirow}
\usepackage{multibib}
\usepackage{caption}
\usepackage{setspace}
\usepackage{breqn}

\emnlpfinalcopy

\def\emnlppaperid{343}

\title{Neural Generation of Regular Expressions from Natural Language\\ with Minimal Domain Knowledge}

\author{Nicholas Locascio \\ MIT \\ njl@mit.edu \And Karthik Narasimhan \\ MIT \\ karthikn@mit.edu \And Eduardo DeLeon \\ MIT \\ edeleon04@mit.edu \AND Nate Kushman \\ Microsoft \\ nate@kushman.org \And Regina Barzilay \\ MIT \\ regina@csail.mit.edu}

\date{\today}

\begin{document}

\maketitle

\begin{abstract}
This paper explores the task of translating natural language queries into regular expressions which embody their meaning. In contrast to prior work, the proposed neural model does not utilize domain-specific crafting, learning to translate directly from a parallel corpus. To fully explore the potential of neural models, we propose a methodology for collecting a large corpus\footnote{Code and data are submitted as supplementary material.} of regular expression, natural language pairs. Our resulting model achieves a performance gain of 19.6\% over previous state-of-the-art models.
\end{abstract}

\section{Introduction}
This paper explores the task of translating natural language text queries into regular expressions which embody their meaning. Regular expressions are built into many application interfaces, yet most users of these applications have difficulty writing them \cite{friedl2002mastering}.  Thus a system for automatically generating regular expressions from natural language would be useful in many contexts. Furthermore, such technologies can ultimately scale to translate into other formal representations, such as program scripts \cite{raza2015compositional}.

Prior work has demonstrated the feasibility of this task. ~\newcite{kushman2013using} proposed a model that learns to perform the task from a parallel corpus of regular expressions and the text descriptions. To account for the given representational disparity between formal regular expressions and natural language, their model utilizes a domain specific component which computes the semantic equivalence between two regular expressions. Since their model relies heavily on this component, it cannot be readily applied to other formal representations where such semantic equivalence calculations are not possible.

In this paper, we reexamine the need for such specialized domain knowledge for this task. Given the same parallel corpus used in ~\newcite{kushman2013using}, we use an LSTM-based sequence to sequence neural network to perform the mapping. Our model does not utilize semantic equivalence in any form, or make any other special assumptions about the formalism. Despite this and the relatively small size of the original dataset (814 examples), our neural model exhibits a small 0.1\% boost in accuracy.

To further explore the power of neural networks, we created a much larger public dataset, \textbf{NL-RX}. Since creation of regular expressions requires specialized knowledge, standard crowd-sourcing methods are not applicable here. Instead, we employ a two-step generate-and-paraphrase procedure that circumvents this problem. During the generate step, we use a small manually-crafted grammar that translates regular expression into natural language. In the paraphrase step, we rely on crowd-sourcing to paraphrase these rigid descriptions into more natural and fluid descriptions. Using this methodology, we have constructed a corpus of 10,000 regular expressions, with corresponding verbalizations.
   
Our results demonstrate that our sequence to sequence model significantly outperforms the domain specific technique on the larger dataset, reaching a gain of 19.6\% over of the state-of-the-art technique.
\section{Related Work}
\paragraph{Regular Expressions from Natural Language}
There have been several attempts at generating regular expressions from textual descriptions. Early research into this task used rule-based techniques to create a natural language interface to regular expression writing \cite{ranta1998multilingual}. Our work, however, is closest to \newcite{kushman2013using}. They learned a semantic parsing translation model from a parallel dataset of natural language and regular expressions. Their model used a regular expression-specific semantic unification technique to disambiguate the meaning of the natural language descriptions. Our method is similar in that we require only description and regex pairs to learn. However, we treat the problem as a direct translation task without applying any domain-specific knowledge.
 
\paragraph{Neural Machine Translation}
Recent advances in neural machine translation (NMT)~\cite{DBLP:journals/corr/BahdanauCB14,devlin2014fast,luong2015effective} using the framework of sequence to sequence learning~\cite{sutskever2014sequence} have demonstrated the effectiveness of deep learning models at capturing and translating language semantics. 

\section{Regex Generation as Translation}
\begin{figure}
  \centering
 \includegraphics[width=1.0\columnwidth]{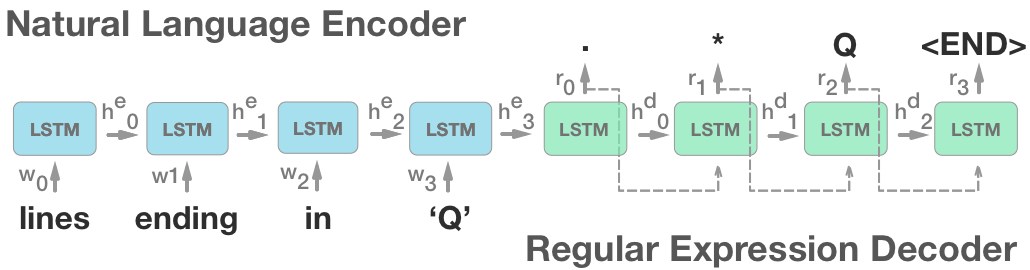}
  \caption{Deep-Regex Encoder-Decoder setup.}
  \label{fig:model}
\end{figure}

Our model is inspired by the recent advancements in sequence to sequence neural translation. We use a Recurrent Neural Network (RNN) with attention~\cite{mnih2014recurrent} for both encoding and decoding (Figure~\ref{fig:model}). 

Let $W = w_1,w_2 ... w_m$ be the input text description where each $w_i$ is a word in the vocabulary. We wish to generate the regex $R = r_1, r_2, ... r_n$ where each $r_i$ is a character in the regex.

We use LSTM~\cite{hochreiter1997long} cells in our model, the transition equations for which can be summarized as:
\begin{align}
\label{eqn:eqlabel}
\begin{split}
	i_t &= \sigma (U^{(i)} x_t + V^{(i)} h_{t-1} + b^{(i)}), \\
	f_t &= \sigma (U^{(f)} x_t + V^{(f)} h_{t-1} + b^{(f)}), \\
	o_t &= \sigma (U^{(o)} x_t + V^{(o)} h_{t-1} + b^{(o)}) \\
	z_t &= \tanh ( U^{(z)} x_t + V^{(z)} h_{t-1} + b^{(z)})	\\
	c_t &= i_t \odot z_t + f_t \odot c_{t-1} \\
	r_t &= o_t \odot \tanh(c_t)
\end{split}
\end{align}
where $\sigma$ represents the sigmoid function and $\odot$ is elementwise multiplication. The input $x_t$ is a word ($w_t$) for the encoder and the previously generated character $r_{t-1}$ for the decoder. The attention mechanism is essentially a `soft' weighting over the encoder's hidden states during decoding:
\begin{dmath*}
\alpha_{t}(e) = \frac{\exp (\text{score}(h_t, h_e))}{\sum_{e'} \exp (\text{score}(h_t, h_{e'}))}
\end{dmath*}
where $h_e$ is a hidden state in the encoder and \emph{score} is the scoring function. We use the general attention matrix weight for our scoring function.

Our model is six layers deep, with one word embedding layer, two encoder layers, two decoder layers, and one dense output layer. Our encoder and decoder layers use a stacked LSTM architecture with a width of 512 nodes. We use a global attention mechanism \cite{DBLP:journals/corr/BahdanauCB14}, which considers all hidden states of the encoder when computing the model's context vector. We perform standard dropout during training \cite{srivastava2014dropout} after every LSTM layer with dropout probability equal to 0.25. We train for 20 epochs, utilizing a minibatch size of 32, and a learning-rate of 1.0. The learning rate is decayed by a factor 0.5 if evaluation perplexity does not increase.

\section{Creating a Large Corpus of Natural Language / Regular Expression Pairs}
Previous work in regular expression generation has used fairly small datasets for training and evaluation. In order to fully utilize the power of neural translation models, we create a new large corpus of regular expression, natural language pairs titled \textbf{NL-RX}.

The challenge in collecting such corpora is that typical crowdsourcing workers do not possess the specialized knowledge to write regular expressions. To solve this, we employ a two-step generate-and-paraphrase procedure to gather our data. This technique is similar to the methods used by ~\newcite{Wang:15} to create a semantic parsing corpus. 

\begin{table}
    \centering
    \caption{Regex $\rightarrow$ Synthetic Grammar for Data Generation}
    \par\medskip
    \resizebox{0.5\textwidth}{!}{%
 \begin{tabular}{| c | c | c |}  
 \hline
 \multicolumn{3}{|c|}{\textbf{Non-Terminals}} \\
 \hline
 x \& y $\rightarrow$ x and y & x $\vert$ y $\rightarrow$ x or y  &  $\sim$(x) $\rightarrow$ not x \\
 \hline
 .*x.*y $\rightarrow$ x followed by y & .*x.* $\rightarrow$ contains x &  x\{N,\} $\rightarrow$ x, N or more times \\
 \hline
 x\& y\& z $\rightarrow$ x and y and z & x $\vert$ y $\vert$ z $\rightarrow$ x or y or z  & x\{1,N\} $\rightarrow$ x, at most N times \\
 \hline
 x.* $\rightarrow$ starts with x & .*x $\rightarrow$ ends with x & \textbackslash b x\textbackslash b $\rightarrow$ words with x \\
 \hline
 (x)+ $\rightarrow$ x, at least once  & (x)* $\rightarrow$ x, zero or more times & x $\rightarrow$ only x \\
 \hline

 \multicolumn{3}{c}{} \\
 \hline
 \multicolumn{3}{|c|}{\textbf{Terminals}} \\
 \hline
 [AEIOUaeiou] $\rightarrow$ a vowel & [0-9] $\rightarrow$ a number  &  word $\rightarrow$ the string `word' \\
 \hline
 [A-Z] $\rightarrow$ an uppercase letter & [a-z]  $\rightarrow$ a lowercase letter & .  $\rightarrow$ a character \\ \hline
 
\end{tabular}%
}%
\label{table:grammar}
\end{table}

\begin{figure}
  \centering
 \includegraphics[width=1.0\columnwidth]{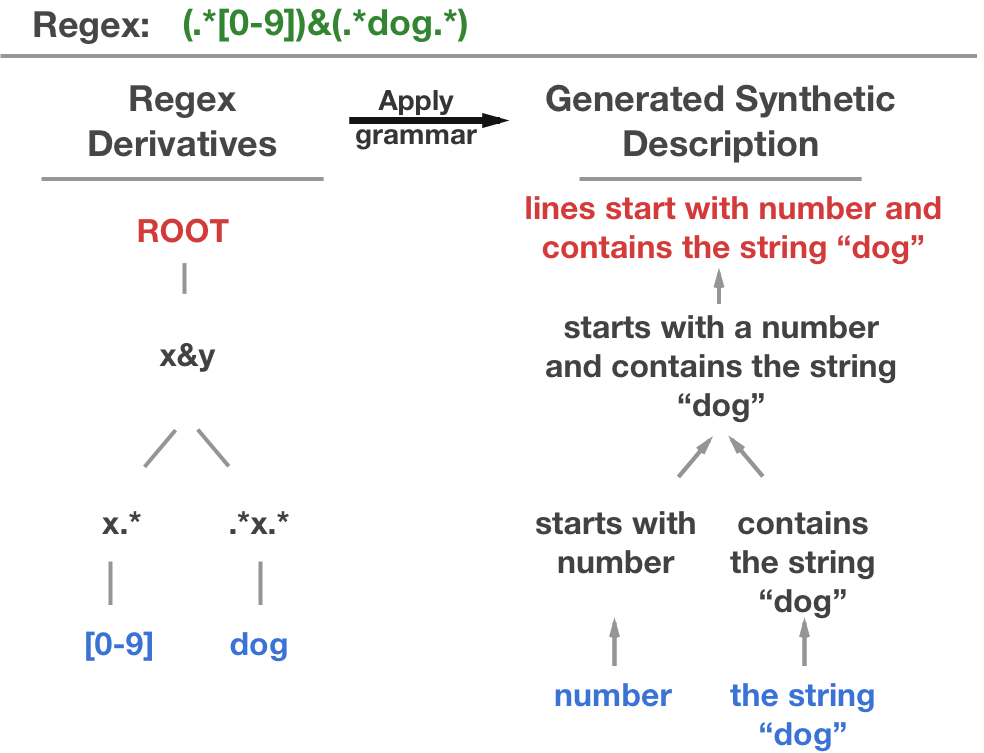}
  \caption{Process for generating Synthetic Descriptions from Regular Expressions. Grammar rules from Table \ref{table:grammar} are applied to a node's children and the resulting string is passed to the node's parent.}
  \label{fig:tree}
\end{figure}

In the generate step, we generate regular expression representations from a small manually-crafted grammar (Table \ref{table:grammar}).  Our grammar includes 15 non-terminal derivations and 6 terminals and of both basic and high-level operations. We identify these via frequency analysis of smaller datasets from previous work \cite{kushman2013using}. Every grammar rule has associated verbalizations for both regular expressions and language descriptions. We use this grammar to stochastically generate regular expressions and their corresponding synthetic language descriptions. This generation process is shown in Figure \ref{fig:tree}.

While the automatically generated descriptions are semantically correct, they do not exhibit richness and variability of human-generated descriptions. To obtain natural language (non-synthetic) descriptions, we perform the paraphrase step. In this step, Mechanical Turk \cite{turk} human workers paraphrase the generated synthetic descriptions into the fluent verbalizations.

\paragraph{NL-RX}
Using the procedure described above, we create a new public dataset (NL-RX) comprising of 10,000 regular expressions and their corresponding natural language descriptions. Table~\ref{table:example} shows an example from our dataset. 

\begin{table}
    \centering
    \caption{NL-RX Text Descriptions and Regular Expression}
    \resizebox{0.5\textwidth}{!}{%
 \begin{tabular}{|r  l|} 
 \hline
 \textbf{Synthetic:} & lines not words with starting with a capital letter \\
 \textbf{Paraphrased:} & lines that do not contain words that begin \\ & with a capital letter \\ 
 \textbf{Regex:} & $\sim($\textbackslash b([A-Z])(.*)\textbackslash b) \\
 \hline
\end{tabular}%
}%
\label{table:example}
\end{table}

Our data collection procedure enables us to create a substantially larger and more varied dataset than previously possible. Employing standard crowd-source workers to paraphrase is more cost-efficient and scalable than employing professional regex programmers, enabling us to create a much larger dataset. Furthermore, our stochastic generation of regular expressions from a grammar results in a more varied dataset because it is not subject to the bias of human workers who, in previous work, wrote many duplicate examples (see Results).

\section{Experiments}
\paragraph{Datasets}
We split the 10,000 regexp and description pairs in NL-RX into 65\% train, 10\% dev, and 25\% test sets.

In addition, we also evaluate our model on the dataset used by \newcite{kushman2013using} (\textsc{KB13}), although it contains far fewer data points (824). We use the 75/25 train/test split used in their work in order directly compare our performance to theirs.

\paragraph{Training}
We perform a hyper-parameter grid-search (on the dev set), to determine our model hyper-parameters: learning-rate $=$ 1.0, encoder-depth $=$ 2, decoder-depth $=$ 2, batch size $=$ 32, dropout $=$ 0.25. We use a Torch ~\cite{Collobert02torch:a} implementation of attention sequence to sequence networks from ~\cite{harvardseq2seq}. We train our models on the train set for 20 epochs, and choose the model with the best average loss on the dev set.

\paragraph{Evaluation Metric}
 To accurately evaluate our model, we perform a \emph{functional} equality check called DFA-Equal. We employ functional equality because there are many ways to write equivalent regular expressions. For example, (a$\vert$b) is functionally equivalent to (b$\vert$a), despite their string representations differing. We report DFA-Equal accuracy as our model's evaluation metric, using \newcite{kushman2013using}'s implementation to directly compare our results.

\paragraph{Baselines}

We compare our model against two baselines:

\emph{BoW-NN:}
BoW-NN is a simple baseline that is a Nearest Neighbor classifier using Bag Of Words representation for each natural language description. For a given test example, it finds the closest cosine-similar neighbor from the training set and uses the regexp from that example for its prediction. 


\emph{Semantic-Unify:}
Our second baseline, Semantic-Unify, is the previous state-of-the-art model from \cite{kushman2013using}, explained above. \footnote{We trained and evaluated Semantic-Unify in consultation with the original authors.}

\section{Results}
\begin{figure}
  \centering
 \includegraphics[width=1.0\columnwidth]{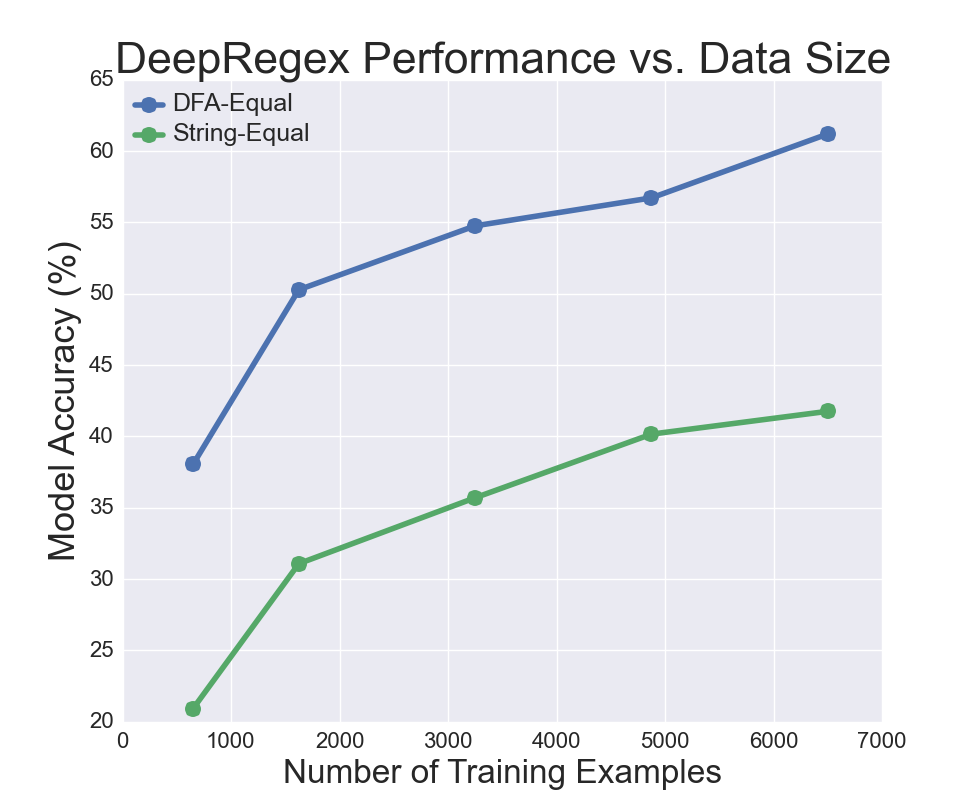}
  \caption{Our model's performance, like many deep learning models, increases significantly with larger datasets. \textbf{String-Equal}:Accuracy on direct string match, \textbf{DFA-Equal}:Accuracy using the DFA-Equal evaluation.}
  \label{acc_vs_data}
\end{figure}

Our model significantly outperforms the baselines on the NL-RX dataset and achieves comparable performance to \emph{Semantic Unify} on the \textsc{KB13} dataset (Table~\ref{table:results}). Despite the small size of \textsc{KB13}, our model achieves state-of-the-art results on this very resource-constrained dataset (814 examples). Using NL-RX, we investigate the impact of training data size on our model's accuracy. Figure \ref{acc_vs_data} shows how our model's performance improves as the number of training examples grows.

\begin{table}
\centering
\resizebox{0.5\textwidth}{!}{%
\begin{tabular}{c | c | c | c | c | c } 
\multirow{2}{*}{\textbf{Models}}  &  \multicolumn{2}{c|}{\textbf{NL-RX-Synth}} & \multicolumn{2}{c|}{\textbf{NL-RX-Turk}} & \textbf{KB13} \\
  & Dev & Test & Dev & Test  & Test \\
  \hline
  BoW NN & 31.7\% & 30.6\% & 18.2\% & 16.4\% & 48.5\% \\ 

 Semantic-Unify & 41.2\% & 46.3\% & 39.5\% & 38.6\% & 65.5\% \\ 

 Deep-RegEx & \textbf{85.75\%} & \textbf{88.7\%} & \textbf{61.2\%} & \textbf{58.2}\% & \textbf{65.6\%} \\
\end{tabular}
}%
\caption{DFA-Equal accuracy on different datasets. \textbf{KB13}: Dataset from Kushman and Barzilay(2013), \textbf{NL-RX-Synth}: NL Dataset with original synthetic descriptions, \textbf{NL-RX-Turk}: NL Dataset with Mechanical Turk paraphrased descriptions. Best scores are in bold.}
\label{table:results}
\end{table}

\paragraph{Differences in Datasets}
Keeping the previous section in mind, a seemingly unusual finding is that the model's accuracy is higher for the smaller dataset, KB13, than for the larger dataset, NL-RX-Turk. On further analysis, we learned that the KB13 dataset is a much less varied and complex dataset than NL-RX-Turk. KB13 contains many duplicates, with only 45\% of its regular expressions being unique. This makes the translation task easier because over half of the correct test predictions will be exact repetitions from the training set. In contrast, NL-RX-Turk does not suffer from this variance problem and contains 97\% unique regular expressions. The relative easiness of the KB13 dataset is further illustrated by the high performance of the Nearest-Neighbor baselines on the KB13 dataset.




\section{Conclusions}
In this paper we demonstrate that generic neural architectures for generating regular expressions outperform customized, heavily engineered models. The results suggest that this technique can be employed to tackle more challenging problems in broader families of formal languages, such as mapping between language description and program scripts. We also have created a large parallel corpus of regular expressions and natural language queries using typical crowd-sourcing workers, which we make available publicly.


\bibliography{emnlp2016}
\bibliographystyle{emnlp2016}

\end{document}